\documentclass[letterpaper, 10 pt, conference]{ieeeconf}  

\IEEEoverridecommandlockouts                              

\usepackage{graphics} 
\usepackage[pdftex]{graphicx}
\usepackage{circuitikz}
\usepackage{tikz}
\UseRawInputEncoding
\usetikzlibrary{calc,patterns,decorations.pathmorphing,decorations.markings,decorations.pathreplacing}
\usepackage{tikzscale}
\usetikzlibrary{patterns,decorations.pathmorphing,positioning}
\usetikzlibrary{arrows}
\usepackage{scalefnt}
\usepackage{subfigure}
\graphicspath{ {./images/} }
\usepackage{amsmath}
\usepackage{float}
\usepackage{circuitikz}
\usepackage{pgfplots}
\usepgfplotslibrary{units}
\usepackage{filecontents}
\usepackage{tikz,filecontents,amsmath}
\usepackage{algorithmic}
\usepackage{amssymb}    

\usepackage{bm,times}
\usepackage{bm} 
\usepackage{color} 
\usepackage{cancel}
\usepackage[export]{adjustbox}
\usepackage{mathtools}
\usepackage{caption}
\usepackage{comment}
\usepackage{fixltx2e}
\usetikzlibrary{calc}

\usepackage{todonotes}
\usepackage{pdfpages}
\usepackage{graphicx}
\usepackage{caption}
\usepackage{booktabs} 
\usepackage{amsmath,bm,mathtools,upgreek}
\usepackage{amssymb,stmaryrd}
\usepackage{prettyref}
\newrefformat{fig}{Figure~[\ref{#1}]}
\newsavebox\IBoxA \newsavebox\IBoxB \newlength\IHeight
\newcommand\TwoFig[6]{
  \sbox\IBoxA{\includegraphics[width=0.3\textwidth]{#1}}
  \sbox\IBoxB{\includegraphics[width=0.3\textwidth]{#4}}%
  \ifdim\ht\IBoxA>\ht\IBoxB
    \setlength\IHeight{\ht\IBoxB}%
  \else\setlength\IHeight{\ht\IBoxA}\fi
  \begin{figure}[!htb]
  \minipage[t]{0.3\textwidth}\centering
  \includegraphics[height=\IHeight]{#1}
  \caption{#2}\label{#3}
  \endminipage\hfill
  \minipage[t]{0.3\textwidth}\centering
  \includegraphics[height=\IHeight]{#4}
  \caption{#5}\label{#6}
  \endminipage 
  \end{figure}%
}
\makeatletter 
\newcommand*{\b@xplus}[1][+]{\ooalign{%
    $\m@th\vcenter{\hbox{$\m@th#1$}}$\cr%
    \hidewidth$\m@th\boxempty$\hidewidth\cr}} 
\renewcommand*{\boxplus}{\mathbin{\b@xplus}} 
\renewcommand*{\boxminus}{\mathbin{\b@xplus[-]}} 
\makeatother

\usepackage{xcolor} %

\usepackage[final]{hyperref}
\usepackage[capitalise]{cleveref}
\hypersetup{linkbordercolor=green}

\title{\LARGE \bf
Kinematically-Decoupled Impedance Control for Fast Object Visual Servoing and Grasping on Quadruped Manipulators
}

\author{Riccardo Parosi$^{1,2,*}$, Mattia Risiglione$^{1,2,*}$, Darwin G. Caldwell$^{1}$, Claudio 
Semini$^{1}$, Victor Barasuol$^{1}$
\thanks{
$^{1}$Dynamic Legged Systems Lab, Istituto Italiano di Tecnologia (IIT), Genova, Italy, \begin{tt}\{name.surname\}@iit.it\end{tt}.
\newline
$^{2}$Dipartimento di Informatica, Bioingegneria, Robotica e Ingegneria dei Sistemi (DIBRIS), Universit\`a di Genova, Genova, Italy,  \begin{tt}\{name.surname\}@edu.unige.it\end{tt}.
\newline
$^{*}$Equal contribution.
}
}

\begin{document}

\maketitle
\thispagestyle{empty}
\pagestyle{empty}


\begin{abstract}
We propose a control pipeline for SAG (Searching, Approaching, and Grasping)
of objects, based on a decoupled arm kinematic chain and impedance control,
which integrates image-based visual servoing (IBVS). The kinematic decoupling
allows for fast end-effector motions and recovery that leads to robust visual
servoing. The whole approach and pipeline can be generalized for any mobile
platform (wheeled or tracked vehicles),  but is most suitable for dynamically moving quadruped manipulators thanks to their reactivity against disturbances. 
The compliance of the impedance controller makes the robot safer for
interactions with humans and the environment. We demonstrate the performance and
robustness of the proposed approach with various experiments on our 140 kg HyQReal quadruped robot
equipped with a 7-DoF manipulator arm. The experiments consider dynamic locomotion,
tracking under external disturbances, and fast motions of the target object. 
\end{abstract}


\section{INTRODUCTION}
To increase the number of tasks mobile manipulation systems can execute in unstructured environments, mobility and vision are two key aspects. Concerning the former, legs allow to select footholds and control the wrench acting on the floating base, orienting and moving it to increase the manipulation workspace when necessary \cite{machines10080719}\cite{https://doi.org/10.48550/arxiv.2210.10044}\cite{Spova}. 
Until now, vision for legged platforms such as quadrupeds and bipeds has been mainly used for locomotion, e.g. to correct nominal footholds \cite{9134750}\cite{8642374} or for navigation, e.g. visual odometry \cite{9852710}. Many recent works combined the advantages of a mobile legged platform with a robotic manipulator to perform manipulation tasks: opening a door \cite{bellicoso2019alma}\cite{act11110304}\cite{sleiman2020MPCplanner}, pulling a rope with a basket \cite{Ferrolho2022RoLoMaRL}, turning a valve \cite{Ferrolho2022RoLoMaRL}\cite{sleiman2020MPCplanner}, grasp a target object \cite{Spova} and put it into a trash bin \cite{https://doi.org/10.48550/arxiv.2210.10044}. Although successful executions, most of these works provide to the robot direct knowledge of its surroundings and do not close the loop with vision for manipulation. 
The use of visual feedback from an onboard camera placed at the arm's end-effector, also known as Eye-In-Hand camera \cite{eyeInHandVseyeToHandCamera}, has been shown to guarantee a more accurate positioning of the arm's end-effector for manipulation, robustness to calibration uncertainties, and reactivity to environmental
changes \cite{538972}. From 2D images position-based visual servoing (PBVS) retrieves the pose of the target, while image-based visual servoing (IBVS) works with feature representation directly in the image domain. The advantages of IBVS over PBVS are the following: \textit{(i)} it does not require any 3D model; 
\textit{(ii)} it is more robust with respect to uncertainties of robot and camera model, in particular to calibration errors \cite{doi:10.1163/156855303322554409};
\textit{(iii)} it is easier to formulate feature-based motion strategies aimed at keeping the target always in the camera field of view. Over the past, IBVS control schemes have been proposed for control of underactuated systems like drones \cite{Jabbari2014AnAS}, non-holonomic mobile robots \cite{1641766}\cite{ALLIBERT200811244} and floating-base space manipulators \cite{ALEPUZ20161}. Despite the advantages, feature depth is unknown in IBVS and it must be estimated or measured (e.g directly from a RGB-D camera) in order to calculate the interaction matrix. 
To alleviate some of the problems induced by both methods, hybrid schemes \cite{760345} use 3D information, usually obtained by epipolar geometry, to control some degrees of freedom (DoF) of the camera, while the remaining ones are controlled through IBVS.

\begin{figure}[t!]
     \centering
        \includegraphics[width=1.0\columnwidth]{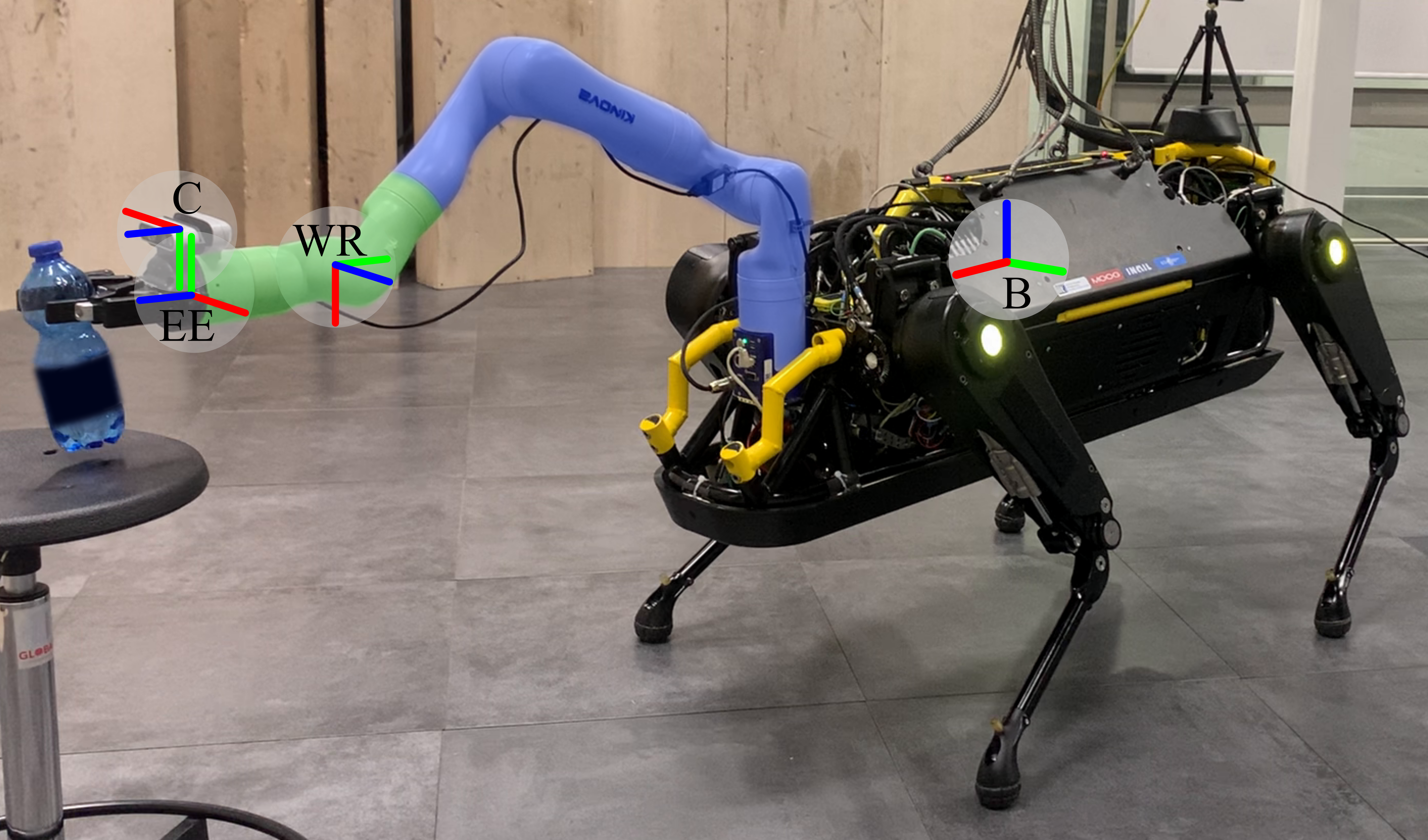}
        \caption{IIT's 140 kg HyQReal robot equipped with a 7-DoF manipulator arm
        (Kinova Gen3 \cite{kinovaGen3}) grasping a bottle. The translucent blue and
        green areas represent, respectively, the Shelbow and the Wrist kinematic
        groups (proposed and introduced in Sec. \ref{arm_controller} to allow for
        fast visual servoing). The frames depicted on the robot associate, respectively,
        the red/green/blue colored axes to their $x/y/z$
        coordinate axes. The frames are named as B: robot's base, WR: arm's wrist, C: camera, EE: arm's end-effector.}
        \vspace{-0.5cm}
        \label{fig: HyQWithKinovaAndBottle}
\end{figure}
To have robots autonomously performing tasks that involve environment interaction, they need to have the ability to grasp objects and/or tools. Depending on the application, e.g. logistics, domestic support, construction, etc., the difference will be in what and how to manipulate. The problem of grasping a generic object is commonly split into three phases: \textit{(a)} Search, i.e. scanning the robot's surroundings, while recognizing and distinguishing objects of interest; \textit{(b)} Approach, i.e. plan and execute a trajectory to get close to the object; \textit{(c)} Grasp, i.e. move the gripper to the target pose in order to grasp the object and close the gripper. 
Previous frameworks have given contributions to sub-problems related to the SAG (S: Search, A: Approach, G: Grasp) of an object. These contributions include ensuring stop-free exploration while searching \cite{8282002}, detecting objects in challenging environments \cite{9093335}, collision-free object search in cluttered scenarios \cite{RFusion} and locating moving targets \cite{inbook}. Additionally, a few-shot object detection is presented in \cite{Griffin_2023_WACV} to learn detection-based tasks for new objects. In \cite{Griffin_2023_WACV}, a complete SAG pipeline is executed, and the robot learns to grasp scattered objects. Although the robot is able to execute the SAG pipeline, the target is not being shown to move.

Mobile platforms are commonly used for SAG problems, due to the increased mobility and reachability. These systems have more degrees of freedom than the ones necessary from the visual task, normally defined as keeping the object in the camera field of view. Hence, the camera position can be changed in many different ways, using redundancy to optimize a configuration-dependent criteria, such as distance from obstacles, singularities and manipulability indices, or dynamic cost functions \cite{DeLuca2007ImagebasedVS}. In \cite{7801041} the authors formulate visual tracking tasks for a humanoid robot in a hierarchical Quadratic Programming (QP) problem, relating the motion of the visual features to the joint accelerations and solving for the latter ones. Other constraints relating feature visibility and mobility, such as dynamic consistency and center of mass control, are set as tasks/constraints in the same QP problem at different priorities. Differently, in \cite{Spova}, vision is not directly used in the motion control of the robot, but only in the motion generation. A 3D target position is retrieved with the camera parameters and fed to a trajectory generator for the quadruped robot Spot with a Kinova robotic arm. The whole-body robot posture is optimized through inverse kinematics to reach the desired position and the motion plan is capable to leverage on the utility of legs to increase the arm reachability and avoid collisions with a table. The platform results to have 80\% accuracy in grasping a ball placed at different initial condition, but during fails the final grasping position results far from the object. The authors relate the problem either to inaccurate ball
position estimation or to possible discrepancy between the real and planned initial robot condition. Additionally, the manipulator is velocity-controlled, as commonly done for visual servoing, making it stiff and not suitable for contexts where human and/or unknown obstacles are in the robot proximity.

Similarly to \cite{Spova}, we tackle the problem of mobile manipulation with a quadruped manipulator and vision in the loop.
More specifically, we propose a control pipeline for the SAG of a target object, using an Eye-In-Hand RGB-D camera mounted at the arm's end-effector. 
In this work, we propose a control approach that uses the joints of the wrist, made up commonly by two or three small and compact actuators, connected through lower inertia links, for the visual task. The rest of the robotic manipulator kinematic chain and the floating base is used to move the robot and position the arm's wrist for Searching, Approaching and Grasping. Based on visual information, we generate a sequence of positions and velocities, converted later to torques for the execution of the SAG sequence. In contrast to the previous mentioned papers, all the actuators are controlled through torque control and an impedance control strategy is integrated to render impedances, through legs, on the quadruped's base and on the arm, through the arm actuators. The impedance rendered at the arm can be chosen to mitigate tracking errors, external disturbance attenuation induced by the floating base and/or by an external source (e.g. manipulation, human interaction).   
The proposed approach is validated with a set of experiments on the 140 kg hydraulic, torque-controlled quadruped HyQReal, with a torque-controlled 7-DoF Kinova Gen3 arm as robotic manipulator. 
In summary, we highlight the following contributions
\begin{itemize}
\item A kinematically-decoupled control approach that integrates
an IBVS scheme and impedance control. The control approach maps
the visual task only on the wrist, exploiting low-inertia links
for fast motion and reactiveness, and the rest of the kinematic
chain for less demanding arm positioning. To the best of the
authors' knowledge, this is the first time visual servoing is
integrated on a fully torque-controlled quadruped manipulator.

\item A sequence of behavior and control signals for the Search, Approach and Grasp with a
torque-controlled quadruped manipulator.

\item Experimental demonstration and assessment of the approach on
a quadruped manipulator, showing active compliance and visual servoing
in presence of external disturbances, the ability to execute the SAG
pipeline, and to track a fast-moving object.

\end{itemize}
The paper is organized as follows: Section \ref{sec:motion_control}
presents the dynamic model and the motion control of the robot. Section
\ref{sec:motion_generation} describes how the whole body motion of the
platform is generated. Section \ref{sec:results} describes the experiments
and discusses the results. Section \ref{sec:conclusions} closes the
paper with conclusions and future work.


\section{Motion Control} \label{sec:motion_control}
\subsection{Robot Model}
The full rigid-body dynamics of a legged manipulator can be described by the set of dynamic equations in \eqref{eq:fullDynamicModel}, where $\bm M$ is the inertia matrix, $\bm {\dot{u}}$ the stacked vector of generalized accelerations, $\bm h$ comprises the gravity, Coriolis and Centrifugal terms, $\bm {\tau}$ the actuation torques. 
The subscripts \textit{b}, \textit{l}, \textit{a}, and \textit{e} stand for base, legs, arm and arm's end-effector, respectively.
The stacked vector of generalized accelerations  $\bm {\dot{u}} = [\bm {\ddot{q}}^T_b, \bm {\ddot{q}}^T_l, \bm {\ddot{q}}^T_a]^T \in \mathbb{R}^{6+n_l+n_a}$ denotes the linear and angular accelerations of the base $\bm {\ddot{q}}_b = [\bm {\ddot{x}}^T_b, \bm {\dot{w}}^T_b]^T \in \mathbb{R}^6$ and the rest of the limb joint accelerations.
 $\bm {F}_g \in \mathbb{R}^{3n_c}$ are the ground reaction forces, where $n_c$ denotes the number of contact feet;  $\bm {F}_e \in \mathbb{R}^3$ denotes the external force acting on the arm's end-effector. $\bm {F}_g$ and  $\bm {F}_e$ are mapped respectively to the base through the contact Jacobians $\bm {J}^T_{st}$ and $\bm {J}^T_{e}$. $\bm {J}^T_{e,a} \in \mathbb{R}^{6\times n_a}$ is the Jacobian matrix from base to end-effector.    \\
\begin{equation}
\begin{split}
\underbrace{
\begin{bmatrix}
\bm M_b & \bm {M}_{bl} & \bm {M}_{ba} \\
\bm {M}_{l b} & \bm {M}_{l} & \bm {M}_{la} \\
\bm {M}_{a b} & \bm {M}_{a l} & \bm {M}_{a} 
\end{bmatrix}}_\text{$\bm {M}$}
\underbrace{
\begin{bmatrix}  
\bm {\ddot{q}}_b \\
\bm {\ddot{q}}_l\\
\bm {\ddot{q}}_a
\end{bmatrix}}_\text{$\bm {\dot{u}}$}
&+
\underbrace{
\begin{bmatrix}
\bm {h}_b \\
\bm {h}_l \\
\bm {h}_a \
\end{bmatrix}}_\text{$\bm {h}$}
 = 
\underbrace{
\begin{bmatrix}
\bm {J}^{T}_{st, b} \\
\bm {J}^{T}_{st, l} \\
\bm {0}_{a \times 3n_c} 
\end{bmatrix}}_\text{$\bm {J}^T_{st}$}\bm {F}_{g} + \\ &+ \underbrace{\begin{bmatrix}
\bm {J}^T_{e,b} \\
\bm {0}_{l \times 3} \\
\bm {J}^T_{e,a} 
\end{bmatrix}}_\text{$\bm {J}^T_{e}$}\bm {F}_e + \underbrace{\begin{bmatrix}
\bm {0}_{6 \times 1} \\
\bm {\tau}_l \\
\bm {\tau}_a 
\end{bmatrix}}_\text{$\bm {\tau}$} 
\end{split}
\label{eq:fullDynamicModel}
\end{equation}

\subsection{Base Controller}
The base of a legged manipulator is commonly considered as the \textit{Trunk}, in which various limbs are connected to. In this work, we use the Trunk Controller proposed in \cite{Focchi2017HighslopeTL}, that imposes a desired wrench on the base, $\bm {W}_b^d \in \mathbb{R}^6$, computed based on position and rotation errors, and it gets mapped to ground reaction forces, $\bm F_g$, considering friction, unilaterality constraints, and force limits of each stance leg \eqref{eq:staticTrunkControllerOpt} as
\begin{align}
\min_{\bm {F}_g} \quad & \left\Vert  \begin{bmatrix}
      \bm I & \cdots & \bm I \\
      [\bm p_{bc_1}]_{\times} & \cdots & [\bm p_{bc_n}]_{\times}
    \end{bmatrix}\bm F_g - \bm {W}_{b}^d \right\Vert ^2_{\bm Q} + \left\Vert \bm {F}_g \right\Vert ^2_{\bm R}\notag \\
\textrm{s.t.} \quad & \bm{\underline{d}} \le \bm C \bm F_g \le \bm {\overline{d}}
\label{eq:staticTrunkControllerOpt}
\end{align}
where $\bm p_{bc} \in \mathbb{R}^{3}$ is the relative distance of the foot in contact with respect to the base, $[ \bm p_{bc}]_{\times}$ denotes the skew-symmetric matrix of vector $\bm p_{bc}$, $\bm C$ is the inequality constraint matrix, $\bm{\underline{d}}$ and $\bm{\overline{d}}$ are lower/upper
bound respectively that ensure that the ground reaction forces lie inside the friction cones and the normal components of the forces are bounded by some user-defined limits. 
The first term in the cost represents the tracking error between the actual and the desired wrench, ${\bm {W}_{b}^d = [{\bm {F}^d_{b}}^T, {\bm T^d_{b}}^T]}^T$, defined as 
\begin{align}
\begin{split}
\bm {F}_{b}^d &= \bm {K}_b(\bm {x}_b^d - \bm {x}_b) + \bm {D}_b(\bm {\dot{x}}^d_b-\dot{\bm{x}}_b)
\end{split} \\
\bm T_{b}^d &= \bm {D}_r(\bm {w}_{b}^d - \bm {w}_b) + \bm {K}_r\bm {e}_r
\end{align}
where $\bm {F}_{b}^d$ and $\bm T_{b}^d$ are respectively the desired force and moment for the base, i.e. ${\bm {W}_{b}^d = [{\bm {F}^d_{b}}^T, {\bm T^d_{b}}^T]}^T$. We define as $\bm {e}_r$ the rotational error, $\bm {D}_r \in \mathbb{R}^{3\times3}$ and $\bm {K}_r \in \mathbb{R}^{3\times3}$ diagonal gain matrices for the derivative and proportional term, respectively. For further implementation details on the friction constraints, we refer to \cite{Focchi2017HighslopeTL}. To guarantee a better motion tracking for the base, we compensate the dynamic coupling effects induced by the legs and arm on the base, using the dynamic model in \eqref{eq:fullDynamicModel}. 

\subsection{Leg Controller}
We compute the torques for each leg by superimposing two control actions: a feedforward term, $\bm {\tau}_{ff}$, obtained mapping $\bm F_g$ from \eqref{eq:staticTrunkControllerOpt} to torques through the stance Jacobian, i.e. $\bm {\tau}_{ff} = -\bm J_{st,l}^T\bm F_g$ ; a feedback term in the form of a PD controller.
The second term is needed to track swing leg trajectories. Hence, the total torque for each leg is computed as
\begin{equation}
\bm {\tau}_l = \bm {\tau}_{ff} + PD(\bm q_l, \dot{\bm q}_l,\bm q_l^d, \dot{\bm q}_l^d)
\end{equation}
For the trajectory generation of the legs we exploit the structure of the Reactive Control Framework \cite{6630926}.  
\subsection{Arm Controller}\label{arm_controller}
The kinematic chain of common manipulators can be split into two groups: what we name as \textit{Shelbow} group, comprising shoulder and elbow joints highlighted with blue in Fig.\ref{fig: HyQWithKinovaAndBottle}, and the wrist, consisting of two or three compact joints highlighted with green in Fig.\ref{fig: HyQWithKinovaAndBottle}. Commercial arms, like \textit{Kinova Gen3} \cite{Kinova} and \textit{Franka-Emika} \cite{FrankaEmika}, have this type of structure, where the last three joints at the wrist have smaller maximum torque peak and are connected through smaller links. The idea of this work is to use the former group to establish an impedance connection with a desired position for the wrist. Throughout the manuscript, we refer to wrist position as the origin of the wrist frame (denoted as WR in Fig. \ref{fig: HyQWithKinovaAndBottle}). For searching, approaching or grasping an object, the desired wrist position is normally defined by the camera and target position. Instead, the wrist is used for tracking an end-effector's trajectory when a target is not in the view of the camera, and to keep the target in the view of the camera once found with the visual feedback received by the Eye-In-Hand camera. 
The Cartesian impedance control imposed at the wrist position is applied using the Shelbow's joints and impedances are rendered in the Horizontal Frame (a reference frame whose xy plane is always
horizontal and its x axis always aligned to the x axis of the robot's base) \cite{6630926}, to reduce cross-coupling effects with the base, as
\begin{align}
\bm {\tau}_{shelbow} &= \bm J_{sh}^T\bigr[\bm {K}^{sh}_p(\bm x^d_{wr} - \bm x_{wr})+\bm {K}^{sh}_d(\bm {\dot{x}}^d_{wr}-\bm {\dot{x}}_{wr})\bigr] +\notag\\ &+ \bm h_{sh}
\label{eq:shelbowImpedanceControl}
\end{align}
where $\bm {h}_{sh}$ is the gravity, Coriolis and Centrifugal torques, $\bm J_{sh} \in \mathbb{R}^{6\times 4}$ is the Shelbow Jacobian matrix obtained extracting the first four columns from $\bm J_{e,a}$, $\bm {K}^{sh}_p \in \mathbb{R}^{3\times3}$ and $\bm {K}^{sh}_d\in \mathbb{R}^{3\times3}$ are virtual springs and dampers gains. Instead, $\bm {x}^d_{wr}$ $\in \mathbb{R}^{3}$ and $\bm{x}_{wr}$ $\in \mathbb{R}^{3}$ are the desired and current Cartesian positions of the wrist, while $\bm {\dot{x}}^d_{wr}$ $\in \mathbb{R}^{3}$ and $\bm {\dot{x}}_{wr}$ $\in \mathbb{R}^{3}$ are the desired and current Cartesian linear velocities of the wrist. Both current and desired positions, as well as velocities, of the wrist are expressed in the Horizontal Frame.
For the wrist, the motion control law is generated according to the given reference. During search of the object, the reference is an end-effector trajectory which is tracked generating the wrist torques as
\begin{align}
\bm {\tau}_{wr} &= \bm J_{wr}^T\bigr[\bm {K}^{wr}_{pc}(\bm x^d_{e} - \bm x_{e})+\bm {K}^{wr}_{dc}(\bm {\dot{x}}^d_{e}-\bm {\dot{x}}_{e})\bigr] + \bm h_{wr}
\label{eq: wristEETracking}
\end{align}
where $\bm J_{wr} \in \mathbb{R}^{6\times 3}$ is the wrist Jacobian matrix obtained extracting the last three columns from $\bm J_{e,a}$, which dependency on the Shelbow joints is omitted. Instead $\bm {K}^{wr}_{pc} \in \mathbb{R}^{3\times3}$ and $\bm {K}^{wr}_{dc}\in \mathbb{R}^{3\times3}$ are virtual springs and dampers gains. The terms $\bm {x}^d_{e}$ $\in \mathbb{R}^{3}$ and $\bm{x}_{e}$ $\in \mathbb{R}^{3}$ are the desired and current Cartesian positions of the end-effector, while $\bm {\dot{x}}^d_{e}$ $\in \mathbb{R}^{3}$ and $\bm {\dot{x}}_{e}$ $\in \mathbb{R}^{3}$ are the desired and current Cartesian linear velocities of the end-effector. Both current and desired positions, as well as velocities, of the end-effector are expressed in the Horizontal Frame.
When vision is activated, e.g. when an object is in the field of view of the camera, then joints' velocities for the wrist are retrieved, and used to obtain setpoints for joints' positions by integration. These joints' position and velocities are tracked to generate the torques for the wrist as
\begin{align}
\bm {\tau}_{wr} &= \bm K^{wr}_{pj}(\bm q^d_{wr} - \bm q_{wr})+\bm {K}^{wr}_{dj}(\bm {\dot{q}}^d_{wr}-\bm {\dot{q}}_{wr})\bigr] + \bm h_{wr}
\label{eq: wristVisualTracking}
\end{align}

\noindent
where $\bm {K}^{wr}_{pj} \in \mathbb{R}^{3\times3}$ and $\bm {K}^{wr}_{dj}\in \mathbb{R}^{3\times3}$ are virtual springs and dampers gains. Instead $\bm q^d_{wr}$ $\in \mathbb{R}^{3}$ and $\bm q_{wr}$ $\in \mathbb{R}^{3}$ are the desired and current wrist joints positions, while $\bm {\dot{q}}^d_{wr}$ $\in \mathbb{R}^{3}$ and $\bm {\dot{q}}_{wr}$ $\in \mathbb{R}^{3}$ are the desired and current wrist joints velocities.

\section{Motion Generation} \label{sec:motion_generation}
In this section, we describe the three main phases of our proposed
method that lead to the grasping of an object: Search, Approach and Grasp.

\subsection{Search:} \label{search}
\begin{figure}[h!]
     \centering
        \includegraphics[width=0.8\columnwidth]{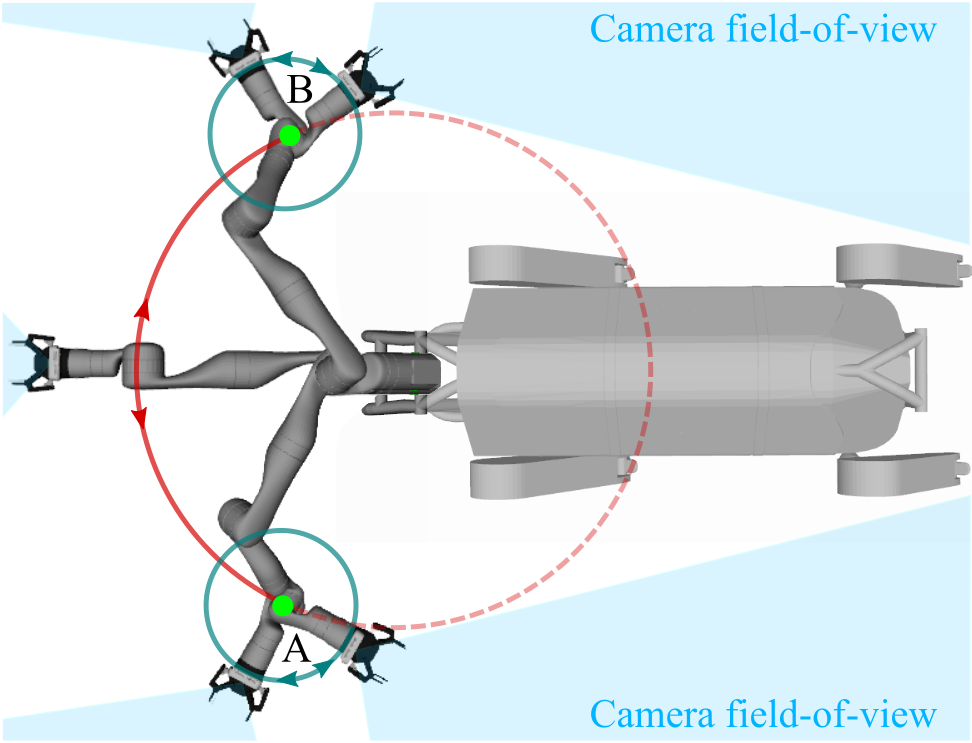}
        \caption{Illustration of HyQReal and its arm at various postures along the
        trajectory paths executed during the object Search phase. The light-green
        dots, A and B, represent the wrist limit positions on its circular path
        around the arm's base. The solid red semi-circle and the green circles
        represent, respectively, the circular searching motion that can be executed
        by the wrist position and the end-effector (the dashed red path is not allowed to avoid
        self-collisions and camera occlusions).}
        \label{fig:hyqreal_search}
\end{figure}

To guide the camera along the search phase, we use an heuristic
trajectory that avoids robot singular configurations and self-collisions.
The trajectory paths for the wrist position and end-effector are illustrated
in Fig. \ref{fig:hyqreal_search}.
First, the arm is brought to a home configuration where the links
are kept away from the base. Then a circular motion, centered around the
arm's base, is tracked by the wrist position using \eqref{eq:shelbowImpedanceControl}
(red path in Fig. \ref{fig:hyqreal_search}). Along this first phase, singularities
are avoided by keeping the radius of this circular trajectory lower than the
wrist maximum allowable distance from the arm's base. To avoid self-collisions
and image occlusions, the scanning behind the quadruped is done in two steps:
first from the left and later from the right of the robot trunk. The two points
that define the limits of the wrist position motion path are indicated in Fig. \ref{fig:hyqreal_search}
as A and B. Once the wrist is in one of these two positions, two circular trajectories
centered around it are used to search backward with the arm's end-effector,
defining $\bm {x}^d_{e}$ and $\bm \dot{{\bm x}}^d_{e}$.
Hence the wrist is rotated, keeping the roll and the pitch
of the end-effector fixed. If the second end-effector backward scan is completed
and no object is detected, the arm returns to the home configuration and the robot is
commanded to rotate around itself by 180 degrees and restart the wrist position and end-effector
search trajectories. We highlight that the only
reason for which the area behind the robot has not been assigned to the arm is to
avoid occlusions with trunk and legs.

\subsection{Approach:} \label{approach}
Once the object has been detected, the robot has to align itself to approach it.
The wrist joints are controlled using the joint impedance controller
\eqref{eq: wristVisualTracking} and for the Shelbow joints the Cartesian impedance
controller \eqref{eq: wristEETracking}. For the alignment, first the wrist is positioned
at the intersection between its circular search trajectory (solid red line in
Fig. \ref{fig:hyqreal_search}) and the line segment connecting the object and the origin
of the robot base frame. If the intersection is located on the dashed part of the red circle,
the wrist position is set at point A or B (depending on the search side).
Successively, the base is aligned to the object by commanding a heading velocity
until the longitudinal axis of the robot aligns with the direction of the object.

In order to keep the object in the camera's field of view during the whole alignment
phase, the references for the controller in \eqref{eq: wristVisualTracking} are given
by the visual servoing. In particular, we consider as features the coordinates of a
pixel in the image projection plane and we define $\bm {s}^*$ as desired value to be the
center pixel in the projection plane, which is (0, 0). The object detection algorithm
outputs the bounding box of the detected object, and we use the coordinates of its center
as current features $\bm s$. We define the feature error as $\bm e = \bm s - \bm s^*$,
which has to be minimized. The interaction matrix \cite{143350} or image Jacobian \cite{4015997},
here referred as $\bm L_s \in \mathbb{R}^{k \times 6}$, where $k$ is the number of features,
links how the features vary if the camera moves. The interaction matrix of a 2D point in the
projection plane is described as follows
\begin{equation*}
\bm L_s = 
\begin{bmatrix}
-\frac{1}{Z} & 0 & \frac{x}{Z} &  xy     & -(1+x^2) & y\\
0 & -\frac{1}{Z} & \frac{y}{Z} & (1+y^2) & -xy      & -x
\end{bmatrix}
\end{equation*}

\noindent where $x$ and $y$ are the coordinates of the point in the projection plane and $Z$ is the $3D$
distance from the camera to the point in Cartesian space. It is common in the literature to refer
to the estimated version of the interaction matrix as $\hat{\bm {L}}_s$, because $Z$ depends
on the camera calibration and on the quantities that are measured.
Using only these features, the end-effector is free to change its roll orientation since all
the rotations around the z axis of the camera are allowed. We impose the camera to stay always
oriented parallel to the base, adding a third feature, $s_{\phi}$,  which constrains such
rotation of the camera. When the camera is oriented parallel to the base, its x axis is always
perpendicular to the z axis of the base frame. Hence, we impose 
\begin{equation}
\bm{x}_c\cdot (\bm R_{cb}\bm{z}_{b})^T = 0  
\label{eq:vs_third_feature}
\end{equation}
where we denote by $\bm x_c$ the camera x axis, $\bm{z}_{b}$ the base z axis, and $\bm R_{cb}$ $\in \mathbb{R}^{3\times3}$ the rotation matrix from base to camera frame. The result of \eqref{eq:vs_third_feature} is $
(\bm R_{cb})_{zx} = 0$,
with $(\bm R_{cb})_{zx}$ being the component at the third row and first column of $\bm R_{cb}$.
The interaction matrix for $s_{\phi}$, can be
derived knowing that the translations of the camera cannot change its orientation, hence the first three columns of $\bm{L}_{s\phi}$ are zero, 
and the time derivative of a rotation matrix is the rotation matrix multiplied by
the skew-symmetric of the angular velocity
\begin{equation}
\bm{L}_{s\phi} = 
\begin{bmatrix}
0 & 0 & 0 & 0 & -(\bm R_{cb})_{zz} & (\bm R_{cb})_{zy}     
\end{bmatrix}
\end{equation}
where $(\bm R_{cb})_{zy}$ and $(\bm R_{cb})_{zz}$ are the components at the second column and third column in the third row of $\bm R_{cb}$. 
Stacking the features, we obtain the following error vector
\begin{equation}
\bm e_s = 
\begin{bmatrix}
x & y & (\mathbf{R}_{cb})_{zx}
\end{bmatrix}^T
\end{equation}
To impose an exponential decay of the error, we derive the twist of the camera expressed in the camera frame as
\begin{equation}
\bm{\xi}^d_{c} = -\lambda\hat{\bm {L}}^{+}_s \bm e_s
\end{equation}
where $\bm L_s \in \mathbb{R}^{3\times6}$ denotes the interaction matrix of the three stacked features and $\hat{\bm L}_s^{+}$ denotes its estimated Moore-Penrose pseudo-inverse. The desired camera twist is mapped to the desired joint velocities and positions for the wrist as
\begin{equation}
  \begin{cases}
    {\bm{\dot{q}}}^d_{wr} =  \bm{J}^{+}_{wr}
    \begin{bmatrix}
        \bm R^T_{cb} & \bm{0} \\
        \bm{0} & \bm R^T_{cb}
    \end{bmatrix}
    \bm{\xi}^d_{c} \\
    q^d_{wr,i} = q^d_{wr,i}(0) + \int_{0}^{T} \dot{q}^d_{wr,i} \,dt , & \forall i = 1,..,N
  \end{cases}
  \label{eq:visual_servoing_reference}
\end{equation}
where $N$ is the number of wrist joints, $q^d_{wr,i}(0)$ the initial desidered joint configuration and $T$ the spanned time interval.

Once the base is aligned with object, the wrist position is kept fixed,
hence the references for \eqref{eq:shelbowImpedanceControl} do not
change. For the end-effector, the wrist is controlled as in \eqref{eq: wristVisualTracking}
using the reference of visual servoing generated by \eqref{eq:visual_servoing_reference}. 
Instead regarding the base, its heading is controlled to be aligned with the one of the arm's end-effector and a walking forward velocity is commanded until the robot reaches the object proximity.
In contrary to the Search,
throughout most of the Approach phase, the robot's base is moving, and
the effects of the legs are indirectly transmitted to the camera through
the arm. We leverage on the capabilities of the impedance controller
implemented at the Shelbow to mitigate tracking and external disturbance
(induced by the base or any other source).

\subsection{Grasp:} \label{grasping}
We enter into the Grasping phase when the robot is in the proximity of the target object and it can compute a grasping pose, i.e. the final position and orientation of the arm's end-effector. During the execution of this phase, the wrist position is kept at the same height of the object, by using the impedance controller \eqref{eq:shelbowImpedanceControl}, and its longitudinal distance is reduced. The latter choice is motivated by the fact that closer the robot moves to the grasping pose, the bigger the object gets in the camera image and the higher the chance for object detection to fail. Instead, the wrist is controlled using \eqref{eq: wristVisualTracking} to keep the object in sight thanks to the reference generated by visual servoing. Subsequently, the robot's base walks to reach the object within the arm workspace. Once that distance is reached, to leverage on the rotational DoFs, the base pitch is commanded to lean down or up, according to the estimated 3D object position as
\begin{equation}
\theta^d_b = \arctan\left(\frac{z_{so}}{x_{so}}\right)    
\end{equation}
where $z_{so}$ and $x_{so}$ denote the relative position of the object with respect to the shoulder, along z and x axis directions of the base frame, respectively. To complete grasping, the Shelbow group and the wrist are commanded to reach the previously computed grasping position. During this last phase, visual servoing is not active anymore being the camera too close to the object, and the grasping is performed open-loop. Reached the grasping pose, the gripper is commanded to close, the base adjusts its pitch, and the arm is repositioned to a default posture. 
The latter arm configuration can be in general optimized for avoiding self-collisions and increasing manipulability.


\section{Results} \label{sec:results}
We performed several experiments to validate the proposed approach
using our 140 kg hydraulic quadruped robot, HyQReal, and a Kinova
Robotic Gen3 arm with 7-DoF. A Realsense D435i \cite{realsenseT435i}
is mounted at the arm's end-effector as shown in Fig. \ref{fig: HyQWithKinovaAndBottle}.
We performed three types of experiments: \textit{a)} disturbances
applied on the arm on both positive and negative y-z axis of base frame while the
robot is performing a trot in place (Section \ref{sec: visual_servoing_plus_disturbance});
\textit{b)} the grasp of a bottle positioned at a fixed, unknown
location in the room and out of the initial camera view (Section \ref{sec: sag});
and \textit{c)} the grasp of a bottle thrown between two humans
(Section \ref{sec: sag_moving_object}).
We used color segmentation throughout all the experiments, as object detection
algorithm, with blue as color to recognize.
The visual servoing gain $\lambda$ is set to 3.0.
For locomotion, a walking-trot gait is used, characterized by the
alternated motion of diagonal leg pairs 
with a step frequency set to 1.3 Hz and a step duty factor equal to 0.6
(the duty factor is the ratio between the time a leg is in stance over
the entire step period). To manage the sequence of actions, we developed
a Behavior Tree based on the open-source project \cite{behaviorTree}
(version 3.8). The Behavior Tree provides reactive behaviors to
unforeseen events, such as the loss of the object from the camera view.
All the experiments described in the following sections are also
included in the accompanying video\footnote{The accompanying video is also available at the following YouTube link: \\\href{https://www.youtube.com/watch?v=ztMl52v3ncY}{https://www.youtube.com/watch?v=ztMl52v3ncY}}. 

\subsection{Visual servoing with external disturbance} 
\label{sec: visual_servoing_plus_disturbance}
\begin{figure}
\centering
\hspace*{-0.03\linewidth}
\vspace{0.0cm}
\begin{tikzpicture}
\node[inner sep=0pt] (plotTrajectoryPositionMode) at (-0.7,0)
    {
    \includegraphics[scale=1,width=1.0\columnwidth]{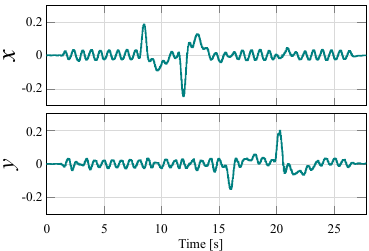}
    };
\end{tikzpicture}
\caption{Response of visual features, x-y center position of detected
object (bottle) in the projection plane. Phases where visual error is higher
coincide to time interval of interaction with human.}
\label{fig: experiment1_visual_servoing_tracking.}
\end{figure}

\begin{figure}
\centering
\hspace*{-0.03\linewidth}
\vspace{0.0cm}
\begin{tikzpicture}
\node[inner sep=0pt] (plotTrajectoryPositionMode) at (-0.7,0)
    {
    \includegraphics[scale=1,width=1.0\columnwidth]{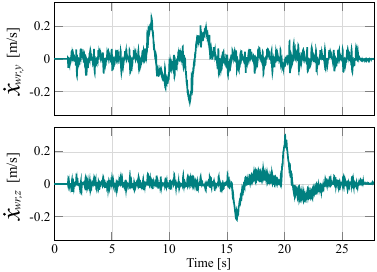}
    };
\end{tikzpicture}
\caption{Translational Cartesian velocities, relative to the Base frame, 
for arm's wrist along y (top) and z axis (bottom). Peaks of
velocities correspond to maximum accelerations caused by human force on the arm.}
\label{fig: experiment1_wrist_velocities.}
\vspace{-0.0cm}
\end{figure}

\begin{figure}[t!]
\centering
\begin{tikzpicture}
\centering
\node (russell) at (0,0)
  {\includegraphics[width=\linewidth]{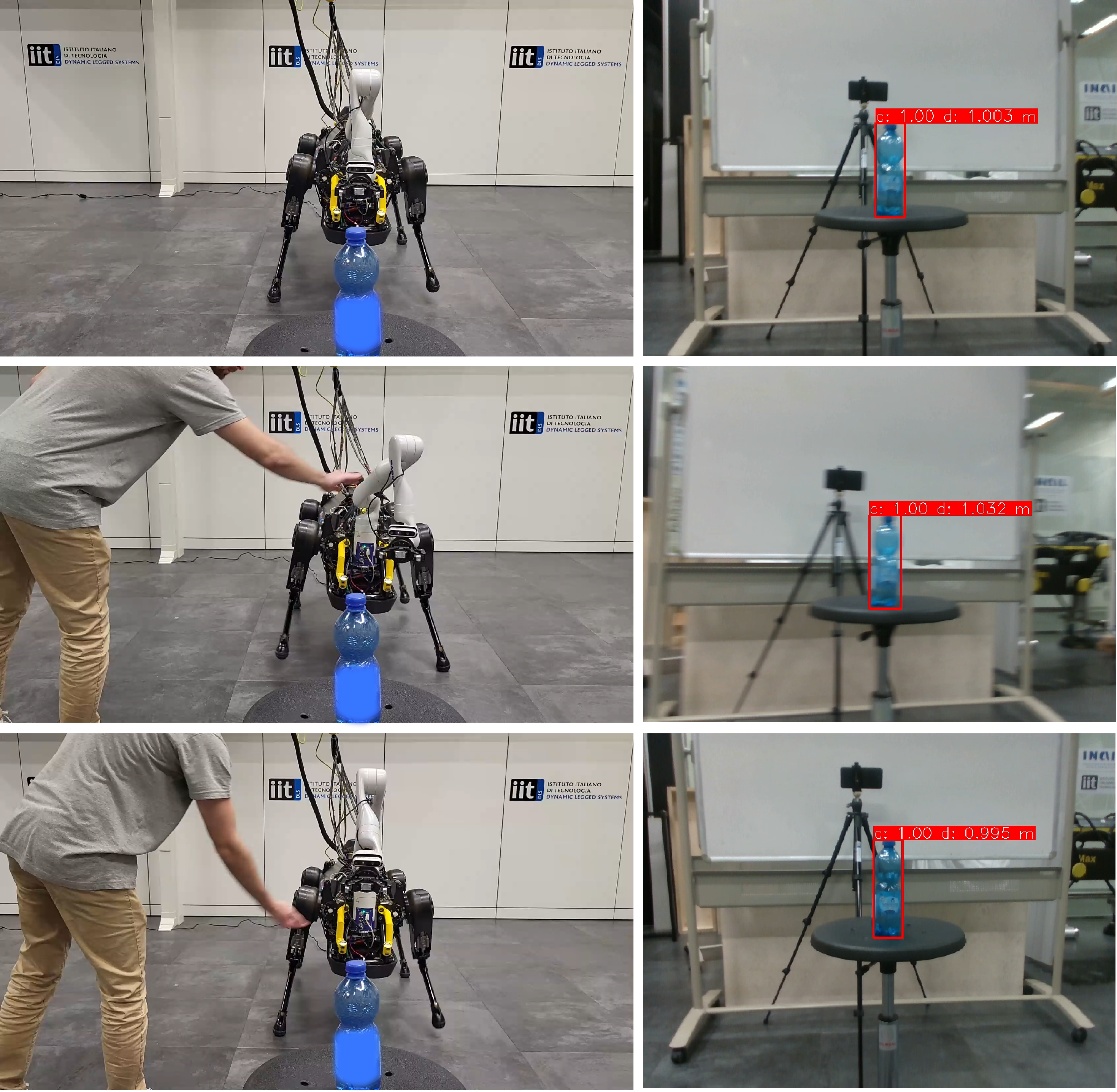}};
  
\end{tikzpicture}
\caption{Recorded sequence of HyQReal and its arm with respect to the object (left column) and the
detected object from the camera view (right column). In the first row, the robot is trotting
in place and, in the middle and bottom rows, the arm's elbow joint is about its maximum displacement from the initial
position along lateral and longitudinal directions, respectively. The displacement is caused by the interaction with a human.}
\label{fig: sequence_disturbaces}
\end{figure}

During this first experiment, the quadruped manipulator is positioned
in front of a bottle at a distance of around 1 m, and is commanded
to keep the object in the field of view of the camera while trotting.
The desired robot behavior is similar to the one expected for the
Approach phase, where the quadruped manipulator has to walk and keep
the object in sight. During the experiment, a disturbance is applied
at the forearm by a human. First, the external forces are applied along
(positive and negative) lateral and vertical directions, respectively
along y and z axis of the robot's base frame. For the arm, the active
controllers are: \eqref{eq:shelbowImpedanceControl} for the Shelbow
group, and \eqref{eq: wristVisualTracking} for the last three joints
of the manipulator. The Cartesian impedances  \eqref{eq:shelbowImpedanceControl}
along y, and z direction are set to: $K^{sh}_p = 50 N/m$, $K^{sh}_d = 5 Ns/m$.
The impedance gains for the joint impedance controller, used by the
last three joints, in \eqref{eq: wristVisualTracking} are set to:
$K^{wr}_{pj} = 100 N/rad$ and $K^{wr}_{dj} = 5 Ns/rad$. When the human applies
the force, the camera view is disturbed, hence the feature tracking
is degraded, as shown in Fig. \ref{fig: experiment1_visual_servoing_tracking.}.
Due to the impedance control strategy applied at the arm's wrist,
the arm responds in a compliant way and does not try to rigidly hold
its position as it would have done under velocity control. Additionally,
the decoupled approach allows to have low impedances at the wrist
because the Shelbow joints are not used for tracking of visual features.
In terms of design, a trade-off can be established between position tracking
accuracy and compliance, according to the context in which the robot
has to operate. From Fig. \ref{fig: experiment1_wrist_velocities.},
when the velocities return to zero, after a disturbance is applied,
the visual features (x-y in projection plane) are zero, i.e. the object
is back centered in the camera, as shown in Fig. \ref{fig: experiment1_visual_servoing_tracking.}
for $t \approx 8s$, $t \approx 12s$, $t \approx 17s$ and $t \approx 21s$. This means by the time the arm
reaches the maximum elongation, the camera still has the object centered,
as shown in the recorded sequence from Fig. \ref{fig: sequence_disturbaces}. 

\subsection{SAG of a bottle}
\label{sec: sag}
\begin{figure}[t!]
\centering
\begin{tikzpicture}
\centering
\node (russell) at (0,0)
  {
  \includegraphics[width=\linewidth]{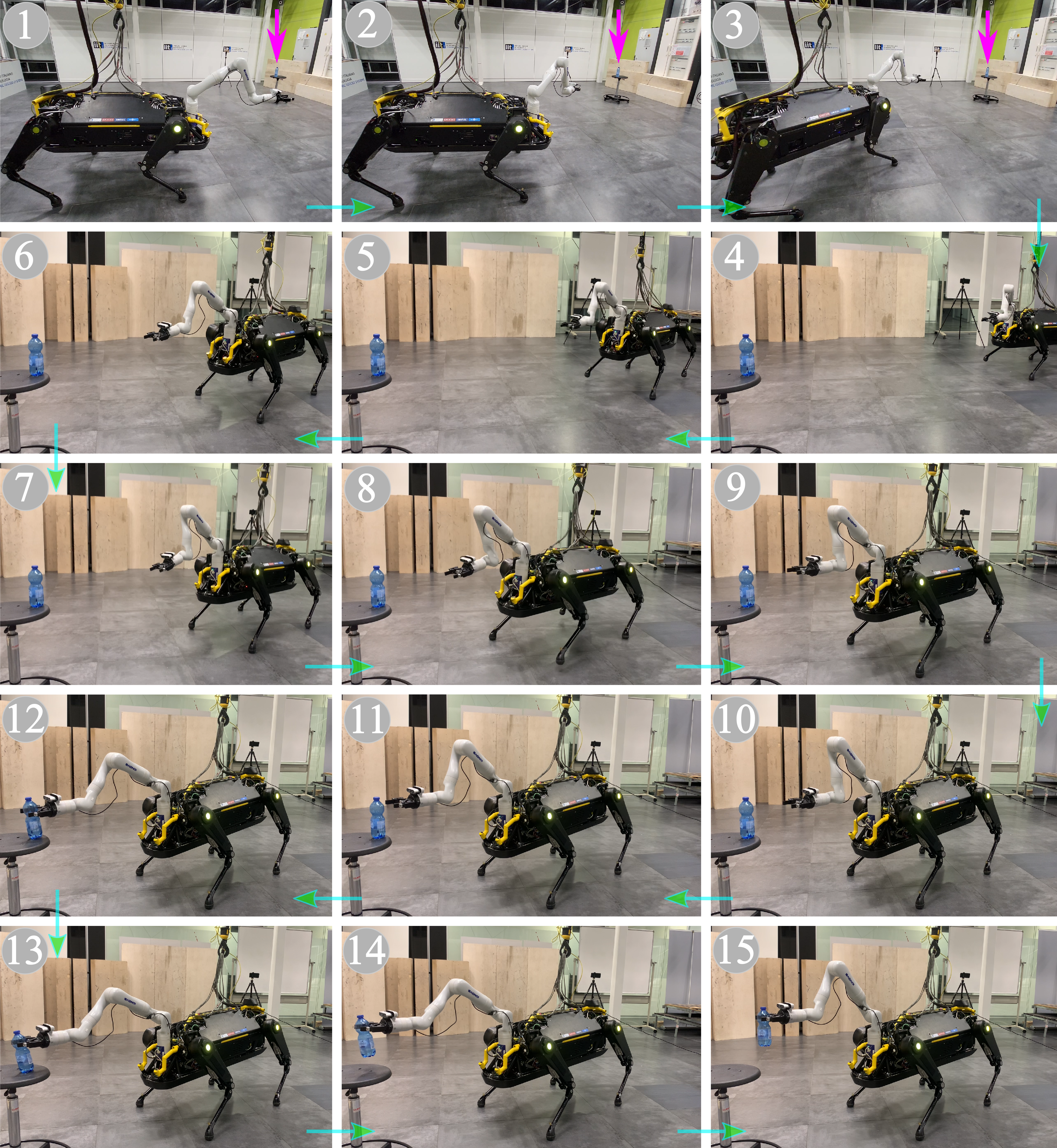}
  };
\end{tikzpicture}
\caption{Recorded sequence of HyQReal with a Kinova arm during the execution of the SAG pipeline. The snapshot
sequence is read from the top to the bottom according to the numbers and light green arrows.
Pink arrows indicate the localization of the bottle for the first three snapshots of the
sequence (i.e., for $S_1$, $S_2$, and $S_3$). At $S_1$ the robot searches for the object. At $S_2$
the object is detected. At $S_3$ the robot adjusts its orientation w.r.t. the object. From $S_4$
to $S_6$ the robot approaches the object performing a walking-trot. From $S_7$ to $S_{15}$
the robot moves its base and arm to prepare and execute the grasping.}
\label{fig: SAGPipeline}
\end{figure}

During this experiment, the quadruped manipulator has to execute the whole SAG pipeline. As it is possible to see from the picture on the left corner of Fig. \ref{fig: SAGPipeline}, the Eye-in-Hand camera is not pointing towards the object at the start. Hence, the robot uses its arm to rotate its camera around the trajectory defined in Section \ref{search}, and it executes the Approach phase defined in Section \ref{approach} after the bottle is detected. The gains used for the active controllers during the Approach phase are defined as the previous experiment \ref{sec: visual_servoing_plus_disturbance}. At the end of the Approach phase, the robot is placed at 0.7 m far from the object. We relied on the Realsense depth estimation measurement to position the robot correctly in front of the object. The grasping phase spans along the last three rows of pictures in Fig. \ref{fig: SAGPipeline}. From the bounding box of the detected object, we retrieve the pixel coordinates of its center, and we use this information, together with the depth measurement and the intrinsic parameters of the camera, to retrieve the 3D object position. As mentioned in Section \ref{grasping}, in the proposed approach, the grasping position is calculated and then the base finalizes its last adjustments to bring the object inside the arm's workspace. Hence, we highlight here that the accuracy of the final grasping position depends on the accuracy of locomotion to properly position the base, and on the state estimation. From trials, we did not experience the need of re-calculating the grasping position during the Grasping phase.

\subsection{Visual tracking of a fast moving bottle}
\label{sec: sag_moving_object}
\begin{figure}[t!]
\centering
\begin{tikzpicture}
\centering
\node (russell) at (0,0)
  {\includegraphics[width=\linewidth]{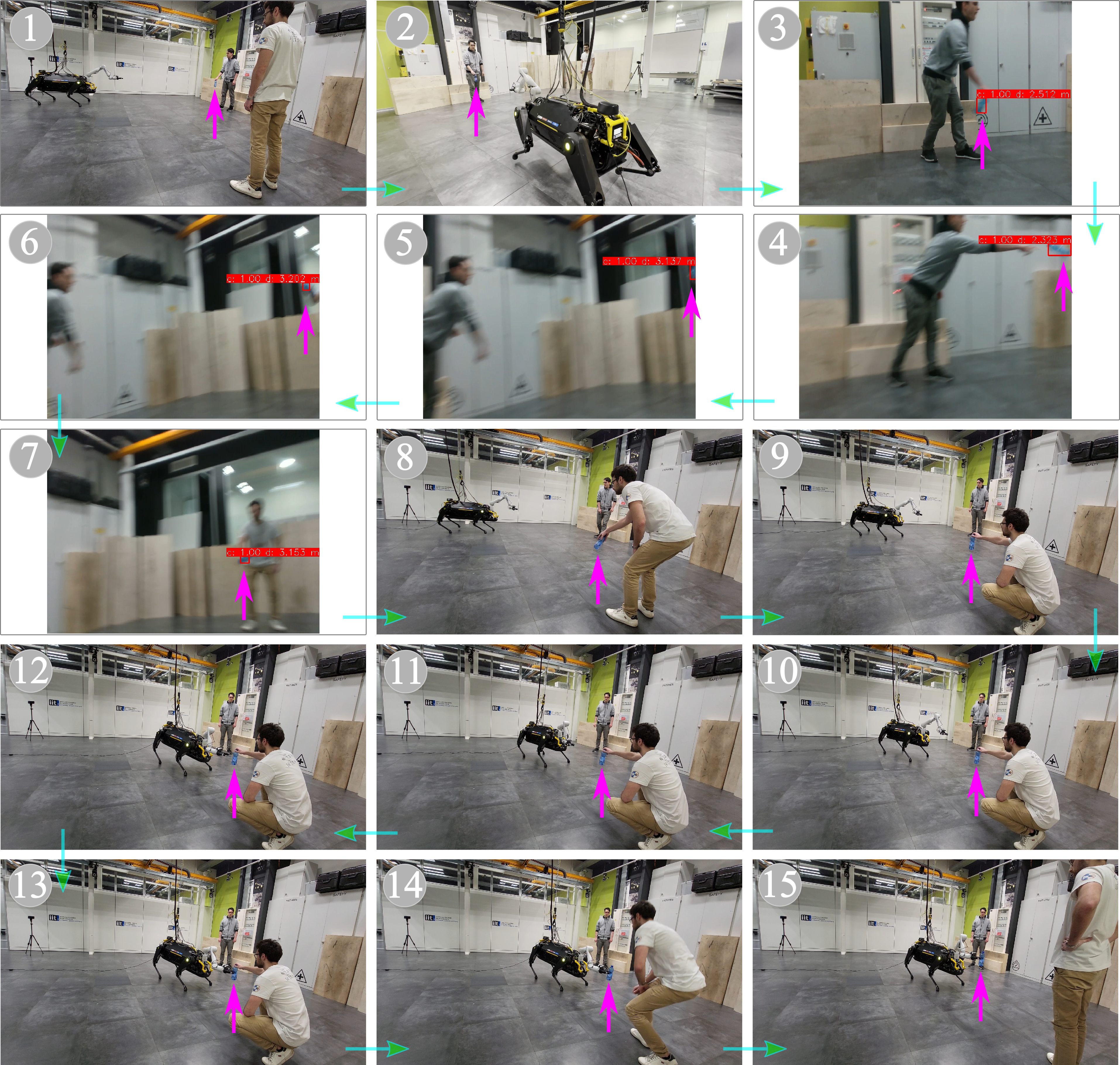}};
\end{tikzpicture}
\caption{Recorded sequence of HyQReal with a Kinova arm during the execution of the SAG
pipeline while performing a walking-trot. The snapshot sequence is read from the top to
the bottom according to the numbers and light green arrows. Pink arrows indicate the
localization of the object in all the snapshots. The target object, handled by the first person,
is detected at the first two snapshots (i.e. $S_1$ and $S_2$). From $S_3$ to $S_7$ the
object is thrown to the second person. The reactiveness of the proposed method allows to
maintain the object tracking from the launching to the catching. From $S_8$ to $S_{10}$
HyQReal approaches the object. The grasping is prepared and executed from $S_{11}$ to $S_{15}$.}
\label{fig: human_throw_exp}
\end{figure}

During this experiment we challenged our system to keep a bottle in sight when throwing
it between two people. More specifically, the quadruped manipulator starts to
execute the SAG pipeline targeting the bottle handled by one of the two people present
in the room. Afterwards, the bottle is thrown and passed from one to the other person.
Along the entire bottle trajectory, the arm's end-effector is able to keep the bottle in the camera
field of view, as shown in the image sequence of Fig. \ref{fig: human_throw_exp}; the error in the visual features drives the camera to point to the
flying object. Thanks to the decoupled approach, the wrist keeps the object centered and it acts as a helm, pointing towards
the direction the base has to walk to. The reactivity of the wrist justifies the choice
and the benefits of mapping directly the visual task to the last three joints of the
manipulator. By executing the SAG pipeline, the quadruped manipulator is driven to grasp
the bottle from the hands of the human as shown in the accompanying video.

\section{Conclusions} \label{sec:conclusions}
In this work, we presented a control pipeline for the Search, Approach and Grasp of an object using a legged manipulator.
The proposed approach defines a behavior sequence for the base and arm to solve the SAG problem, which integrates IBVS to maintain the object in the field of view of the camera and impedance control to render an active compliant behavior on the base and at the level of the wrist position. The main idea of the paper relies on assigning the visual task to the wrist, comprised normally by the last two or three joints, and the rest of the kinematic chain to place the wrist  position in space.
To validate the control approach, we executed experiments where the robot uses visual servoing and gets disturbed. The results show the arm's compliance and its ability to keep the object centered, thanks to the fast motions of the wrist. Additionally, we executed the complete SAG pipeline for grasping a bottle standing on a stool and to track and grasp a bottle thrown between two humans.
As future work we aim at dealing with more complex surroundings, by considering obstacles and exploiting the robot positioning and posture to improve manipulation tasks. Additionally, in order to deploy this control architecture in more complicated and crowded environments, more robust neural network architectures are needed to properly detect and segment objects. 

\bibliographystyle{./bibtex/IEEEtran} 
\bibliography{root}
\addtolength{\textheight}{-12cm}   
\end{document}